\documentclass[english,12pt,draftclsnofoot,onecolumn]{IEEEtran}

\usepackage[utf8]{inputenc} 
\usepackage[T1]{fontenc}    
\usepackage{wrapfig}
\usepackage{hyperref}       
\usepackage{url}            
\usepackage{booktabs}       
\usepackage{amsfonts}       
\usepackage{nicefrac}       
\usepackage{microtype}      

\usepackage{multirow}
\usepackage[textsize=tiny]{todonotes}
\presetkeys{todonotes}{color=blue!20}{}
\usepackage{graphicx,graphics,subcaption}
\usepackage{tikz}
\usetikzlibrary{shapes.geometric, arrows}
\usepackage{pgfplots}

\usepackage{gensymb}

\title{Rotate the ReLU to implicitly sparsify deep networks}

\author{
  Nancy Nayak$^{*}$\thanks{*Corresponding author. \texttt{Email: ee17d408@smail.iitm.ac.in}}, Sheetal Kalyani$^{**}$ \thanks{** \texttt{Email: skalyani@ee.iitm.ac.in}} \\
  Department of Electrical Engineering\\
  Indian Institute of Technology Madras\\
  Chennai, India.\\
}

\begin{document}

\maketitle

\begin{abstract}
  In the era of Deep Neural Network based solutions for a variety of real-life tasks, having a compact and energy-efficient deployable model has become fairly important. Most of the existing deep architectures use Rectifier Linear Unit (ReLU) activation. In this paper, we propose a novel idea of rotating the ReLU activation to give one more degree of freedom to the architecture. We show that this activation wherein the rotation is learned via training results in the elimination of those parameters/filters in the network which are not important for the task. In other words, rotated ReLU seems to be doing implicit sparsification. The slopes of the rotated ReLU activations act as coarse feature extractors and unnecessary features can be eliminated before retraining. Our studies indicate that features always choose to pass through a lesser number of filters in architectures such as ResNet and its variants. Hence, by rotating the ReLU, the weights or the filters that are not necessary are automatically identified and can be dropped thus giving rise to significant savings in memory and computation. Furthermore, in some cases, we also notice that along with saving in memory and computation we also obtain improvement over the reported performance of the corresponding baseline work in the popular datasets such as MNIST, CIFAR-10, CIFAR-100, and SVHN. 
\end{abstract}
\section{Introduction}
Machine Learning has gained a lot of attention recently for surpassing human-level performance in solving problems starting from real-life applications to complex tasks. It leverages the function approximation capabilities of deep networks i.e. a deeper network has more ability for approximation and therefore better accuracy. To achieve a better performance, most of the time a bulkier model is chosen with the hope that more parameters will lead to architecture with more representative power. For example, for the task of image classification, convolutional neural network (CNN) based deep networks like AlexNet \cite{krizhevsky2012imagenet} and VGGNet\cite{simonyan2014very} were proposed that have $60$ Mn and $138$ Mn parameters respectively. With very deep architectures, even if the number of parameters is high, the performance need not improve proportionally. With the architectures like DenseNet\cite{huang2017densely} and GoogleNet\cite{szegedy2015going} new benchmark accuracies were achieved with a lesser number of parameters than AlexNet and VGGNet. Note that the bigger networks with great representation power have a downside - with more layers, these networks are prone to overfitting and deeper networks frequently encounter exploding and vanishing gradient problems. Therefore to ease the training of the network, \cite{he2016deep} proposed a residue-based framework that is much deeper than VGGNet but still has lesser computational complexity. It can be scaled up to thousands of layers as well but for each fraction of a percent of improvement, one needs to double the number of layers. So \cite{zagoruyko2016wide} proposed WideResNet (WRN) that improves the performance by widening the ResNet architectures instead of increasing depth. Blindly increasing depth to increase the approximation power of deep networks may lead to storing many unnecessary parameters for each layer.

Reducing complexity without degrading the performance has become an area of increasing importance in the age of energy-efficient green algorithms. With the increasing number of applications that make use of deep learning (DL) based solutions, deploying a massive DL model at the resource-constrained user-end is challenging. In many application areas such as wireless communication, the devices are typically small i.e. have both limited memory and battery life. Different compression techniques like pruning and quantization, transfer learning, etc. have been introduced to reduce complexity and thereby make it suitable to deploy at the edge network \cite{han2015deep}. Different regularization techniques \cite{han2015learning} are used to sparsify and prune network parameters. However if the sparsification is not done in a structured way, it can only reduce the number of nonzero parameters, and therefore the pruning will not reduce computation even though it saves memory \cite{wen2016learning}. To achieve structural sparsity, Group LASSO was proposed in the context of statistical learning theory for structured sparse vectors where the weights are grouped and an entire group of weights is forced to go to zero by a regularized loss function \cite{meier2008group}. Recently Group LASSO has been used in the context of pruning neural networks as well because the elements of a complete row/column of a weight matrix can be grouped together and can be dropped if not necessary for solving the task at hand\cite{wang2017novel,nayak2020green}. However, this method has multiple hyperparameters and is therefore difficult to tune. In this work, our first aim is to structurally sparsify the architecture. For this, we neither use separate regularizers nor do we threshold the parameters directly. As most of the DL architectures use Rectified Linear Unit (ReLU)\cite{nair2010rectified} and its variants like leaky ReLU \cite{maas2013rectifier}, PReLU\cite{he2015delving} and randomized leaky ReLU \cite{srivastava2014dropout} as the activation functions, we propose an alternate compression technique based on a simple and novel concept of the rotation of ReLU activation. Note that, when we coin the term rotated ReLU, we are only rotating the linear part of the activation function. We discuss the proposed rotated ReLU (RReLU) in detail in Sec. \ref{sec:RReLU}.

To study the capability of the proposed technique of rotated ReLU, we have chosen mainly the ResNet based architectures as they are powerful feature extractors and are extensively used in computer vision, and natural language processing problems and in diverse areas such as security, healthcare, remote sensing, wireless communication, etc. \cite{guo2020gluoncv,nguyen2021secure, alrabeiah2020millimeter, manas2021seasonal,shankar2021binarized}. The question we attempt to answer is, \textit{do we really need such deep/wide networks with millions of parameters for solving a task?} If not, then how to remove or reduce these parameters or filters in a structured fashion such that both memory and computation get saved with very negligible/no reduction in performance. Irrespective of the architecture, any model that has ReLU can be retrained by replacing the ReLUs with RReLUs. This makes the RReLU in our opinion, so powerful tool for neural network compression.

\paragraph{Contributions:}
We propose the rotated ReLU activation and show that it is a powerful tool to save memory and computing power in any architecture. Next, we have shown that for simple tasks like CIFAR-10, the ResNet-56 trained from scratch with RReLU achieves $8.23\%$ saving in memory and $17.28\%$ saving in computation with an increase in performance by $0.68\%$. In the case of deeper architectures, training with RReLU sometimes may not converge to the optimal point as the training is sensitive toward the value of the rotation. For those cases, the weights/filters are initialized with parameters of a pre-trained ReLU network, and the rotation parameters are initialized with $1$. This version achieves nearly equal accuracy for ResNet-110 when compared with the vanilla ResNet-110 for CIFAR-100 with $22.09\%$ and $35.20\%$ gain in memory and computation respectively. As RReLU learns to dropout, RReLU in the WRN-$40$-$4$ architecture helps to save $60.81\%$ in memory requirement and $74.16\%$ in computation with a small improvement in the performance of $0.67\%$ due to better generalization. The proposed RReLU based architectures seem to be implicitly sparsifying the network. We shall see in the later sections that the sparsification is also structured and hence leads to saving not only in memory but also in computation. We have also shown how RReLU works as a coarse feature extractor and with careful pruning of the unnecessary features this reduces the complexity even further. We have validated the results with extensive simulation using fully connect, ResNet, and WideResNet architectures and MNIST, CIFAR-10, CIFAR-100, and SVHN datasets.

\section{Why and how to rotate the ReLU?}
\label{sec:RReLU}
A single layer of a neural network (NN) of depth $L$ is given by $\mathbf{h}_{l+1}=\sigma(\mathcal{F}(\mathbf{h}_l; \mathbf{W}_l))$, where $\mathcal{F}$ is either the weights multiplication operation for fully connected NN (FCNN) or the convolution operation for a convolutional NN (CNN), $\mathbf{W}_l$ are the trainable weights (/filters) of layer $l$ and $\sigma$ is the non-linearity or the activation function. Out of many non-linearities existing today, ReLU\cite{glorot2011deep} ($ReLU(x)=\max(0, x)$) is very popular \cite{cybenko1989approximation, barron1993universal} since it is able to overcome the vanishing gradient problem. It takes a composite of piecewise linear approximation \cite{huang2020relu} and hence can approximate a wide range of functions \cite{hanin2019universal}. It allows nonnegative features to pass as it is and stops the negative features. While approximating any function with the help of ReLU, the slope of the linear part of ReLU is fixed to $1$. In this paper, we propose to increase the degree of freedom of ReLU by rotating it and thereby increase the representative power of ReLU. The rotations or the slope of the ReLUs are trainable i.e. learned during training. Due to the freedom in choosing the rotation, the architecture with RReLU needs lesser trainable weights ($\mathbf{W}_l$) to achieve a similar performance as ReLU. Note that with sufficient trainable weights/filters, both ReLU and RReLU have similar approximation capability and if an architecture with ReLU has enough trainable weights/filters ($\mathbf{W}_l$), the rotations in RReLU can be absorbed by $\mathbf{W}_l$. But interestingly, with RReLU, the slopes are trained in such a way that the weights/filters go to zero in a structured way. A major issue with the deep network is that with increasing depth, the gain one achieves diminishes. For example, on CIFAR-10, to go from accuracy of $93.03$ to $93.57$ using ResNet one has to increase the depth from $56$ to $110$. This now leads to a huge increase in the number of parameters i.e. to gain a $0.54\%$ improvement in accuracy, the number of parameters go up from $0.85$ Mn to $1.727$ Mn. RReLU helps to achieve a similar or better accuracy with a huge saving in memory and computation. By applying the simple but clever method of rotation, multiple vectors in the weight matrix and multiple matrices in the filters can be ignored leading to a significant reduction in the number of parameters and computation in fairly deep networks.

\begin{figure}[t]
    \centering
    \begin{minipage}{0.46\textwidth}
    \begin{subfigure}{0.49\linewidth}
    \resizebox{\linewidth}{!}{
    \begin{tikzpicture}
    \begin{axis}[
        xlabel={\Large x},
        ylabel={\Large y},
        xmin=-1, xmax=1,
        ymin=-1, ymax=1,
        legend pos=north east,
        legend style={nodes={scale=1.3, transform shape}},
        xmajorgrids=true,
        ymajorgrids=true,
        grid style=dashed,
    ]
    \addplot[thick,red,domain=-1:1, thick] {max(0,3*x)};
    \addlegendentry{\(a_l=+1, b_l\geq0\)}
    \addplot[thick,dotted,blue,domain=-1:1, thick] {-max(0,2*x)};
    \addlegendentry{\(a_l=+1, b_l<0\)}
    \addplot[thick,green,domain=-1:1, thick] {max(0,-1*x)};
    \addlegendentry{\(a=-1, b_l\geq0\)}
    \addplot[thick,dotted,magenta,domain=-1:1, thick] {-max(0,-2.5*x)};
    \addlegendentry{\(a=-1, b_l<0\)}
    \end{axis}
    \end{tikzpicture}
    }
    \caption{Rotated ReLU}
    \label{fig:act}
        \end{subfigure}%
        \begin{subfigure}{0.49\linewidth}
        \resizebox{\linewidth}{!}{
        \begin{tikzpicture}
    \begin{axis}[
        xlabel={\Large x},
        ylabel={\Large y},
        xmin=-1, xmax=1,
        ymin=-1, ymax=1,
        legend pos=north west,
        legend style={nodes={scale=1.3, transform shape}},
        xmajorgrids=true,
        ymajorgrids=true,
        grid style=dashed,
    ]
    \addplot[thick,red,domain=-1:1, thick] {max(0,3*x)};
    \addlegendentry{\(b_l\geq0\)}
    \addplot[thick,dotted,blue,domain=-1:1, thick] {-max(0,2*x)};
    \addlegendentry{\(b_l<0\)}
    \end{axis}
    \end{tikzpicture}
    }
    \caption{Unique RReLU}
    \label{fig:finalRReLU}
    \end{subfigure}
    \begin{subfigure}{0.46\linewidth}
        \includegraphics[scale=0.47]{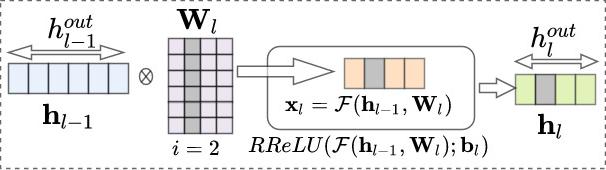}
        \caption{FCNN}
        \label{fig:FCNN}
    \end{subfigure}
    \end{minipage}
    \begin{minipage}{0.49\textwidth}
    \begin{subfigure}{0.49\linewidth}
        \includegraphics[scale=0.42]{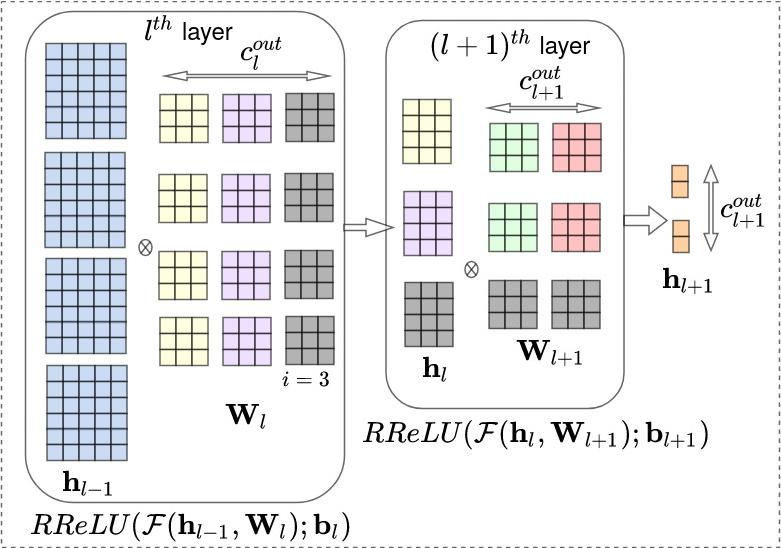}
        \caption{CNN}
        \label{fig:cnn}
    \end{subfigure}
    \end{minipage}

    \caption{\textbf{(a)} and \textbf{(b):} Proposed activation function \textbf{(c) FCNN:} RReLU applied at $l^{th}$ layer of a FCNN. Here $\otimes$ denotes matrix multiplication and $\mathbf{x}_l=\mathbf{h}_{l-1}\otimes\mathbf{W}_l$. In this example, let the slopes $\mathbf{b}_l$ of RReLU are trained in such a way that the $2^{nd}$ element of $\mathbf{b}_l$ is very small i.e. $b_l^{\{2\}}\approx 0$. So the $2^{nd}$ entry in $\mathbf{h}_l$ is negligible as the RReLU does not allow to pass the second element of $\mathbf{x}_l$, clearly the $2^{nd}$ column of $\mathbf{W}_l$ can be ignored (in grey). \textbf{(d) CNN:} RReLU applied to $l^{th}$ convolution layer allows to free up filters that are not necessary for the task. Here $\otimes$ denotes convolution operation. In this example at $l^{th}$ layer, $4$ two dimensional features $\mathbf{h}_{l-1}$ are convolved with $3$ set of filters denoted by $\mathbf{W}_l$ with $4$ sub-filters each, resulting in $\mathbf{h}_l$. The first 2D feature (yellow) of $\mathbf{h}_l$ is calculated by convolving each of the $4$ 2D features with corresponding sub-filters of the first filter (yellow) and adding them. The same occurs at layer $l+1$. Now if due to rotation, let the slopes $b_l^{\{3\}}\approx 0$, then $3^{rd}$ feature in $\mathbf{h}_l$ is also close to zero. In this case the $3^{rd}$ filter of $\mathbf{W}_l$ and the $3^{rd}$ sub-filter of every filter in $\mathbf{W}_{l+1}$ can be ignored. The parameters those can be ignored are highlighted in grey.}
\end{figure}
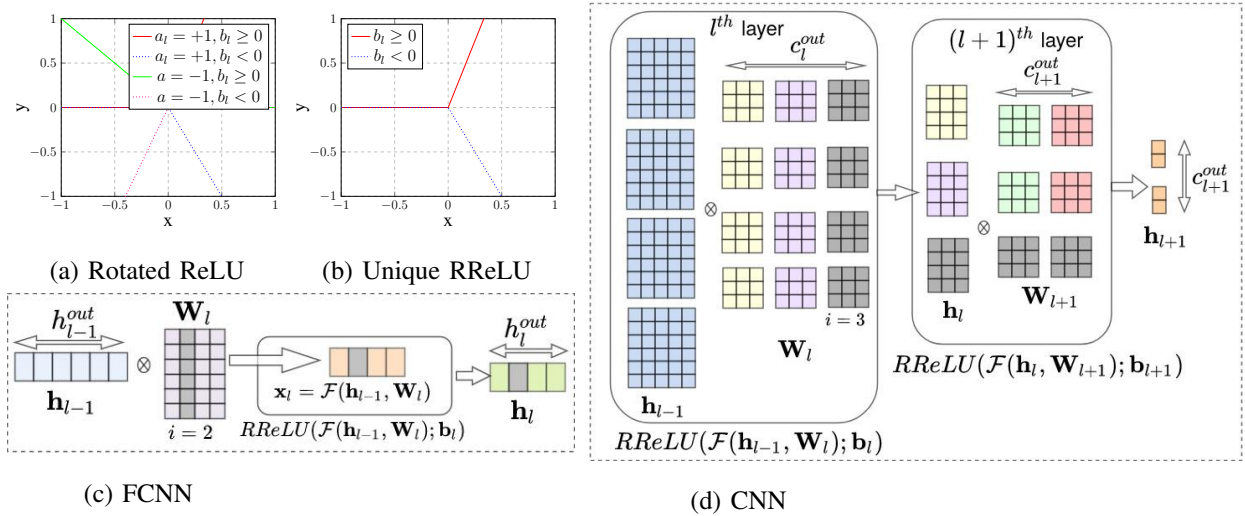

The RReLU activation is given by:
\begin{equation}
\label{tRReLU}
    RReLU(x;a,b)=b\max(0, ax)
\end{equation}
where the trainable parameters are $a\in\{+1,-1\}$ and $b\in \mathbb{R}$. For either the positive or the negative part of the domain, RReLU gives an output of zero, whereas, for the other part of the domain, it is linear with a variable slope depending on $(a,b)$. With these values of $a$ and $b$, all possible RReLUs can be realized and are shown in Fig. \ref{fig:act}. Note that using a single RReLU with a pair of learned $(a,b)$ for all the features across all the layers is very similar to simple ReLU where both $a$ and $b$ are adjusted by the network weights/filters. Therefore to exploit the power of rotation, RReLUs with different pair of $(a,b)$ is used for different features at each layer. Therefore, the feature $\mathbf{h}_l$ at $l^{th}$ layer is given by:
\begin{equation}
    \mathbf{h}_l = RReLU(\mathbf{x}_l;\mathbf{a}_l,\mathbf{b}_l)=\mathbf{b}_l\max(0, \mathbf{a}_l.\mathbf{x}_l)
\end{equation}
where $\mathbf{x}_l=\mathcal{F}(\mathbf{h}_{l-1},\mathbf{W}_{l})$. In RReLU, the rotation of the linear part controls how strongly the convoluted feature $\mathbf{x}_l$ will pass via $l^{th}$ layer. So if the value of $b_l^{\{i\}}$ for some $i\in\mathcal{I}$ is comparatively less then it implies that the convoluted features $x_l^i, \forall i \in\mathcal{I}$ at $l^{th}$ layer are not important for the task. The training with RReLU takes more time to converge because of additional parameters (one for every feature at each layer). But once fully trained, based on the values of $\mathbf{b}_l$, some features can be fully ignored keeping the performance intact. Note that the slopes $\mathbf{b}_l$ converge to different values including $1$. Using RReLU, the representation capability of the resultant piece-wise function increases therefore the number of parameters needed to approximate the same function decreases. As the feature to $l^{th}$ layer before passing it via RReLU is given by $\mathbf{x}_l=\mathcal{F}(\mathbf{h}_{l-1};\mathbf{W}_{l})$, it is clear that $\mathbf{a}_l\in\{+1,-1\}$ can be easily adjusted by the weights/filters $\mathbf{W}_l$. Therefore the only two unique types of RReLU that are sufficient to be used as the activations are given by 
\begin{equation}
    \mathbf{h}_l = RReLU(\mathbf{x}_l;\mathbf{b}_l)=\mathbf{b}_l\max(0, \mathbf{x}_l)
\end{equation}
and are shown in Fig. \ref{fig:finalRReLU}. Note that the value of $h_l^{\{i\}}$ is insignificant when $b_l^{\{i\}}$ is approximately equal to zero and this is logically consistent as $\mathbf{x}_l$ is batch normalized before passing it via RReLU and therefore cannot take a value greater than $1$.  In the next section we describe how rotation helps to achieve lower complexity with lesser computation but still allows one to use deeper networks. For this study, we discuss the results using different architectures based on either fully connected neural network whose basic operation is matrix multiplication or ResNets whose basic operation is convolution and skip connection.

\paragraph{RReLU in FCNN}
Given that the input and output for $l^{th}$ layer is $\mathbf{h}_{l-1}$ and $\mathbf{h}_{l}$ respectively, and the number of elements in $\mathbf{h}_{l}$ is $h_l^{out}$, the number of different RReLUs with trainable $\mathbf{b}_l\in \mathbb{R}^{h^{out}_l}$ is equal to the number of neurons in $l^{th}$ layer which is equal to $h_l^{out}$. After training, element $i,\forall i\in\mathcal{I}$ of $\mathbf{b}_l$ may take values close to zero which shows that the $i^{th}$ feature at $l^{th}$ layer is not useful. This implies that for FCNN, the $i^{th}$ column of the weight matrix $\mathbf{W}_l$ can be ignored. This is pictorially illustrated with an example in Fig. \ref{fig:FCNN}. After training, the value of $b_l^{\{i\}}$ is compared against a threshold $\gamma$, and all the values of $b_l^{\{i\}}$ that are less than $\gamma$ is made zero and the corresponding column of the weight matrix are ignored such that the performance of the network does not degrade. By making a complete row/column zero, the rotation facilitates the structural sparsity for free without any tradeoff in the performance. 

\paragraph{RReLU in CNN}
Given that the number of channel out from the convolution operation at $l^{th}$ layer is $c^{out}_l$, we use $c^{out}_l$ different RReLUs with trainable $\mathbf{b}_l\in \mathbb{R}^{c^{out}_l}$ for $c^{out}_l$ convolution outputs.
If after training, the element $i,\forall i\in\mathcal{I}$ of $\mathbf{b}_l$ takes a value close to zero, then the $i^{th}$ feature at $l^{th}$ layer is not useful. Therefore at $l^{th}$ layer, $i^{th}$ filter out of a set of $c^{out}_l$ filters can be ignored and at $(l+1)^{th}$ layer, $i^{th}$ sub-filter from a set of $c^{out}_l$ sub-filters of each of $c^{out}_{l+1}$ filters can be ignored. This is pictorially explained with an example in Fig. \ref{fig:cnn}. In CNN, all $i\in\mathcal{I}$ filters for which of $b_l^{\{i\}}$ is lesser than a threshold $\gamma$ can be dropped without any negative effect on the performance. In our proposed method, we achieve structural sparsity implicitly without any regularization and pruning methods. Hence we can provide a performance that is close to or sometimes better than the original network, unlike regularized or pruning-based methods. The slopes force the set of rows/columns of the weight matrices to zero in the case of FCNN and force the filters to go to zero in the case of CNN. In the next section, we discuss the saving in memory and computation for both FCNN and CNN using the proposed method.

\section{Saving in memory and computation using RReLU}
\label{sec:saving}
Finding the saving in memory and computation for a FCNN is straightforward. Considering the length of the input and output features of layer $l$ as $h_{l-1}^{out}$ and $h_{l}^{out}$ respectively, and $\mathbf{W}_l\in\mathbb{R}^{h_{l-1}^{out}\times h_{l}^{out}}$, the total number of parameters at $l^{th}$ layer is given by $h_{l-1}^{out}h_{l}^{out}$. During forward pass, number of multiplications and addition required are given by $h_{l-1}^{out}h_{l}^{out}$ and $(h_{l-1}^{out}-1)h_{l}^{out}$ respectively; therefore approximately a total of $2h_{l-1}^{out}h_{l}^{out}$ FLOPs are needed. If $n$ values of $\mathbf{b}_l$ are close to zero, then $n$ columns of $\mathbf{W}_l$ can be ignored. So the memory saving at $l^{th}$ and $(l+1)^{th}$ layer are $h_{l-1}^{out}n$ and $nh_{l+1}^{out}$ respectively. The savings in FLOPs for the $l^{th}$ and $(l+1)^{th}$ layer are approximately $2h_{l-1}^{out}n$ and $2nh_{l+1}^{out}$. 

The savings are even more for the convolution based ResNet architectures. Before that let us first discuss the number of parameters and FLOPs in a single convolution layer. The memory and FLOPs for $1$D-CNN and $2$D-CNN were given in \cite{vikas} and \cite{shankar2021binarized} respectively. Considering $L$ layers in a ResNet architecture, the trainable parameters are given by $\phi = \{\mathbf{W}_1, \dots, \mathbf{W}_L\}$ where $\mathbf{W}_l \in \mathbb{R}^{c_{out}^l \times c_{in}^l \times k \times k}$ is the filter for the $l^{th}$ layer of a 2D CNN. Here $c_{out}^l$ and $c_{in}^l$ represent the input and output channels respectively and $k$ is the dimension of the filter. Note that $c_{out}^l$ features are input to $(l+1)^{th}$ layer therefore $c_{in}^{l+1}=c_{out}^l$. The input and output for the $l^{th}$ layer are $\mathbf{a}_l \in \mathbb{R}^{c_{in}^l \times h_{in}^{lw} \times h_{in}^{lh}}$ and $\mathbf{a}_{l+1} \in \mathbb{R}^{c_{out}^l \times h_{out}^{lw} \times h_{out}^{lh}}$ respectively. Here, $(h_{in}^{lw}$, $h_{in}^{lh})$ and $h_{out}^{lw}$, $h_{out}^{lh}$ are spatial dimensions (width, height) of the input and the output respectively. The total number of multiplication for $l^{th}$ layer is $c_{in}^l \times k^2 \times h^{lw}_{out}\times h^{lh}_{out} \times c_{out}^l$ and the total number of addition for $l^{t h}$ layer is $\left(c_{in}^l-1\right) \times(k^2-1) \times h^{lw}_{out} \times h^{lh}_{out} \times c_{out}^l$. The total count of FLOPs for $l^{th}$ layer of a real-valued 2D CNN is the summation of the number of multiplication and addition that is roughly twice the number of multiplication given by $2 \times c_{in}^l \times k^2 \times h^{lw}_{out}\times h^{lh}_{out} \times c_{out}^l$. Because of the addition of residue in the ResNet structure, if there is a residual connection at $l^{th}$ layer, an extra of $c_{out}^l\times h^{lw}_{out}\times h^{lh}_{out}$ additions are required. The RReLU makes some of the slopes values insignificant therefore the corresponding output and input channels can be ignored. For example, if the input to RReLU at $l^{th}$ layer has $c_{out}^l$ channels and $n$ entries of $\mathbf{b}_l$ are insignificant, then only $(c_{out}^l-n)$ channels remain significant. This needs to save only $(c_{out}^l-n)c_{in}^lk^2$ parameters for $l^{th}$ layer. The computation for these reduced $(c_{out}^l-n)$ filters at $l^{th}$ layer is only $2c_{in}^l k^2 h^{lw}_{out} h^{lh}_{out}(c_{out}^l-n)$ and the same at $(l+1)^{th}$ layer is given by $2(c_{out}^l-n)k^2h^{(l+1)w}_{out}h^{(l+1)h}_{out}c_{out}^{(l+1)}$.

\section{Results and discussion}
\label{sec:results}
To establish the efficacy of the proposed method, we investigated it with different architectures such as a fully connected dense network, different ResNet architectures with depths of $20$, $56$, and $110$ layers, and WideResNet with depths of $40$ and $16$ and widening factor $4$. A wide range of image classification tasks is considered based on increasing difficulty levels starting from MNIST and SVHN to CIFAR-10, CIFAR-100.
In the forthcoming subsections, we have shown the performance of the architectures with the proposed RReLU activation. The potential of RReLU for compactification is also demonstrated in terms of saving memory and computing power. The baseline performances are reproduced with the help of well tuned set of codes for both ResNet and WideResNets.\footnote{Links: \url{https://github.com/akamaster/pytorch_resnet_cifar10}, \url{https://github.com/xternalz/WideResNet-pytorch}} The codes with RReLU for different architectures and the corresponding savings in memory and computation are uploaded as supplementary material. We have used single NVIDIA-GeForce 2080 Ti GPU for all of the experiments.

MNIST is one of the simplest datasets with $60$k training samples and $10$k test samples whereas both CIFAR-10 and CIFAR-100 datasets have $50$k training and $10$k test images and $10$ and $100$ classes respectively. So CIFAR-100 has fewer samples per class compared to CIFAR-10 and hence the classification task using CIFAR-100 is comparatively a difficult task. Another dataset SVHN consists of cropped images of digits similar to MNIST. But unlike MNIST, SVHN has more natural scene images. The top row of Table \ref{tab:RReLUperf} perform the baseline ReLU architectures, followed by the number of parameters and FLOP counts at the second and third row respectively. The training process depends on how the slopes are initialized. The experimental results to show how RReLU helps with different initialization of the architecture are discussed below.

\subsection{Initialization of the architecture to enable maximum saving of resources}
If the slopes are initialized randomly with Gaussian distribution, many of the slopes go to very less value long before convergence which reduces the representation power of the network to a great extent. So it is important to have a good initialization for the slopes such that the network gets enough time to converge before some of the slopes go to zero. In this regard, we propose two types of parameter initialization:

\subsubsection{Type I: Initialize $\mathbf{b}_l$ with Truncated Gaussian Mixture Model}
\label{sec:type-I}
In this setup, the architecture with RReLU with both the weights/filters and the ResNet-20 are trained from scratch. Here the weights/filters are initialized with Kaiming He initialization \cite{he2015delving} and the slopes $\mathbf{b}_l, \forall l\in L$ are initialized with a truncated Gaussian Mixture Model with mean $\{+1,-1\}$ and variance $3$ as shown in Fig. \ref{fig:gmm} in the Appendix. Such a range of values for $\mathbf{b}_l$ gives the network more freedom to arrive at a different optimal point with some of the slopes trained to very less value resulting in a lesser number of parameters. Usually, the deeper networks are generally sensitive toward hyperparameters and therefore the convergence of the training process heavily depends on different hyperparameters such as the number of epochs the network is trained for, the learning rate, the scheduler, etc. The deeper networks with RReLU also are sensitive towards the slopes $\mathbf{b}_l$ of RReLU along with these hyperparameters. Whereas the shallower architectures use most of the weights/filters in every layer for a task, the deeper architectures need not use so many filters per layer to solve the same task. Therefore in the case of the deeper architectures, all the weights/filters that are not necessary for the given task get discarded by the proposed RReLU activation and a huge saving in memory and computation can be obtained.

The total number of trainable parameters is slightly more in architecture with RReLU before pruning compared to architecture with ReLU because the trainable parameters of architecture with simple ReLU are only the weights/filters whereas the same for architecture with RReLU are the rotation parameters or the slopes given by $\mathbf{b}_l, \forall l\in L$ along with the weights/filters. Also, the optimization process is sensitive towards $\mathbf{b}_l$ as a little change in $\mathbf{b}_l$ can take the optimizer to a point far from the current state on the loss surface. So the architectures with RReLU take more time to converge and therefore are trained for more time. Where the standard architectures are trained for $200$ epochs, the networks with RReLU are trained for $1200$ epochs using either multistep learning rate (MLR) scheduler or cosine annealing (CA) scheduler. Note that, different learning rates and different schedulers may vary the performance and this work is not a mere improvement of performance rather it shows how using RReLU can improve the saving in complexity without any negative effect on performance. In the fourth row of Table \ref{tab:RReLUperf}, the percentage of either the total number of columns of weight matrices (FCNN) or the number of filters (CNN) that are ignored is presented. Because of this, the corresponding saving in memory and FLOPs are shown in the fifth and sixth rows. Accuracy after discarding the unnecessary weights/filters is listed in the seventh row of Table \ref{tab:RReLUperf}.

For the comprehensive study of RReLU, in the first experiment, we take a fully connected neural network (FCNN) that has a hidden layer with $500$ neurons. Both the architectures with ReLU and RReLU are trained for $300$ epochs with the same Adam optimizer and have a similar range of accuracy.  Because of using RReLU, the slopes get trained to a range of values that are compared with a threshold $\gamma$, and all the slopes that are lesser than the $\gamma$ can be dropped. The $\gamma$ can be set during deployment in such a way that dropping the weights or filters does not degrade the performance. Please see Table \ref{tab:RReLUperfgamma} in the Appendix for different values of $\gamma$ chosen for deployment. Here $24$ out of $500$ RReLU slopes ($4.8\%$ of all the filters) are less than $\gamma=1.0$ and could be dropped that essentially making the $24$ out of $500$ columns of the weight matrix at the hidden layer unnecessary. The number of parameters of the architecture with ReLU is $0.1$ Mn whereas the architecture with RReLU after discarding the unnecessary weights has $0.097$ Mn parameters keeping the accuracy close to the architecture with ReLU. The number of FLOPs reduces to $0.378$ Mn compared to $0.4$ Mn in ReLU. The savings are even more when the architectures are either deep or have more parameters due to convolution operation. 

\begin{table}[t]
    \caption{Performance of RReLU  in terms of accuracy, number of trainable parameters and computation power (in FLOPs) when trained from scratch. The number of parameters and number of FLOPs are in million (Mn).}
    \label{tab:RReLUperf}
    \centering
    \footnotesize
    \begin{tabular}{p{3.2cm}p{0.6cm}p{0.7cm}p{0.7cm}p{0.7cm}p{0.9cm}p{0.7cm}p{0.7cm}p{0.7cm}p{0.9cm}p{0.7cm}}
    \toprule
    Dataset &\multicolumn{1}{c}{MNIST} &\multicolumn{4}{c}{CIFAR-10} & \multicolumn{4}{c}{CIFAR-100} &\multicolumn{1}{c}{SVHN}\\
    \cmidrule(lr){1-1} \cmidrule(lr){2-2} \cmidrule(lr){3-6} \cmidrule(lr){7-10} \cmidrule(lr){11-11} 
        Architecture & FCNN & ResNet-20 & ResNet-56 & ResNet-110 & WRN-$40$-$4$ & ResNet-20 & ResNet-56 & ResNet-110 & WRN-$40$-$4$ & WRN-16-4\\
        \midrule
        Acc ReLU & $98.33$ & $91.25$& $93.03$ & $93.57$ & $95.47$ & $68.20$ & $69.99$& $74.84$ & $78.82$ & $97.01$\\
        $\#$Params ReLU & \textcolor{red}{$0.10$}& \textcolor{red}{$0.27$} & \textcolor{red}{$0.85$} & \textcolor{red}{$1.72$} & \textcolor{red}{$9.27$} & \textcolor{red}{$0.27$} & \textcolor{red}{$0.85$} & \textcolor{red}{$1.72$} & \textcolor{red}{$9.29$} & \textcolor{red}{$3.08$}\\
        FLOPs ReLU & \textcolor{red}{$0.40$} & \textcolor{red}{$81$} & \textcolor{red}{$231$} & \textcolor{red}{$504$} & \textcolor{red}{$2776$} & \textcolor{red}{$81$} & \textcolor{red}{$231$} & \textcolor{red}{$503$} & \textcolor{red}{$2776$} & \textcolor{red}{$982$}\\
        \midrule
        Filters ignored ($\%$) & $4.8$ & $2.23$ & $8.48$ & $35.66$ & $44.04$ & $0.0$ & $15.33$ & $39.36$ &  $18.30$ & $20.88$\\
        $\#$Params RReLU & \textcolor{blue}{$0.09$} & \textcolor{blue}{$0.26$} & \textcolor{blue}{$0.78$} & \textcolor{blue}{$1.13$} & \textcolor{blue}{$3.19$} & \textcolor{blue}{$0.27$} & \textcolor{blue}{$0.798$} & \textcolor{blue}{$0.876$} & \textcolor{blue}{$6.79$} & \textcolor{blue}{$2.33$}\\
        FLOPs RReLU & \textcolor{blue}{$0.38$} & \textcolor{blue}{$79$} & \textcolor{blue}{$191$} & \textcolor{blue}{$249$} & \textcolor{blue}{$779$} & \textcolor{blue}{$81$} & \textcolor{blue}{$175$} & \textcolor{blue}{$175$} & \textcolor{blue}{$1363$} & \textcolor{blue}{$635$}\\
        Acc RReLU post-pruning & $98.24$ & $93.12$ & $93.71$ & $91.44$ & $95.73$  & 
        $67.66$ & $68.96$ & $57.27$ & $80.72$ & $97.06$\\
        \bottomrule
    \end{tabular}
    
\end{table}

Towards this, we have considered ResNet architectures whose main operation is convolution. The complexity of the network depends on the number of filters used in every layer which is a hyperparameter. Till now there was no way to find out the unnecessary filters and remove them without affecting the performance. By using RReLU, the network is trained to predict using as less filters as possible. Even if the depth of the architecture is increased, in case certain filters are not necessary, they are dropped by introducing RReLU to the architecture. In terms of performance, the architecture with RReLU on ResNet-56 with CIFAR-10 achieves an accuracy of $93.74$ which is better than the reported accuracy of $93.57$ for ResNet-110. The proposed RReLU method declares $171$ out of a total of $2016$ filters ($8.48\%$ of the total filters) unnecessary by training the corresponding $171$ slopes to a value below $\gamma=0.06$ and the number of parameters reduces from $0.85$ Mn to $0.78$ Mn and number of FLOPs reduce from $231.47$ Mn to $191.47$ Mn. Therefore RReLU gives a saving of $8.23\%$ in memory and $17.28\%$ in computation. Note that $\gamma$ is not just one more hyperparameter for training but it can be set using cross validation to some value during testing so that due to the ignored filters the performance is not degraded. Even though sometimes the performance of different ResNet architectures with RReLU is better than one with ReLU, training very deep architecture with RReLU from scratch (i.e. filters are initialized by Kaiming He initialization \cite{he2015delving}) might be difficult as seen from Table \ref{tab:RReLUperf}. Because for the deeper networks like ResNet-110, the training is sensitive towards different training hyperparameters like training epoch, a learning rate scheduler, etc. For ResNet with RReLU, the slopes are added on top of these hyperparameters and make the training even more challenging. Therefore for the deeper architectures, a smarter training method that leads to better results than training from scratch is discussed in Sec. \ref{sec:smarterinit}. However, the strength of the proposed method does not lie in the performance but lies in the fact that without any additional compression technique, the proposed methods learn which filters or which weights are not necessary for training. For example, ResNet-110 architecture with RReLU does not perform as well as one with ReLU (a drop of $2.13\%$ in accuracy) but it can drop the memory requirement by $34.30\%$ and computation by a huge $50.53\%$ margin for CIFAR-10. Interestingly, for a not so deeper but a wider architecture like WRN-$40$-$4$ whose depth is $40$ and widening factor is $4$ and has $9.27$ Mn trainable parameters, the RReLU method saves $65.52\%$ in memory and $71.94\%$ in computation for CIFAR-10. The proposed RReLU is tested with CIFAR-100 and SVHN as well and the results are given in Table \ref{tab:RReLUperf}. 

\subsubsection{Type II: Initialize with a pretrained architecture with ReLU}
\label{sec:smarterinit}
\begin{figure}[t]
    \centering
    \begin{subfigure}{0.16\linewidth}
    \includegraphics[scale=0.28]{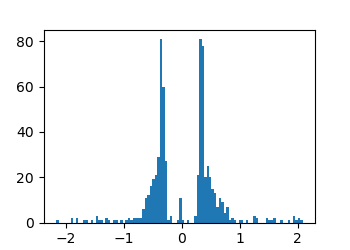}
    \caption{X:20, Y:10}
    \label{fig:init1a}
    \end{subfigure}%
    \begin{subfigure}{0.16\linewidth}
    \includegraphics[scale=0.28]{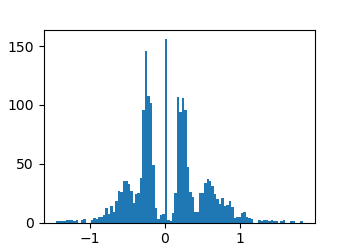}
    \caption{X:56, Y:10}
    \label{fig:init1b}
    \end{subfigure}%
    \begin{subfigure}{0.16\linewidth}
    \includegraphics[scale=0.28]{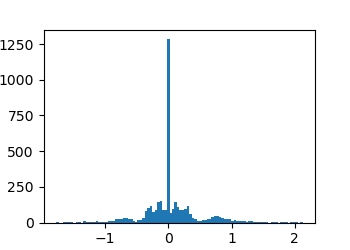}
    \caption{X:110, Y:10}
    \label{fig:init1c}
    \end{subfigure}%
    \begin{subfigure}{0.16\linewidth}
    \includegraphics[scale=0.28]{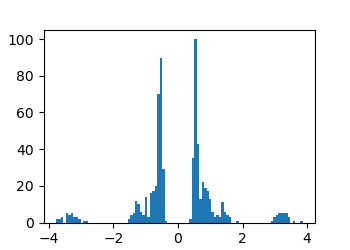}
    \caption{X:20, Y:100}
    \label{fig:init1d}
    \end{subfigure}%
    \begin{subfigure}{0.16\linewidth}
    \includegraphics[scale=0.28]{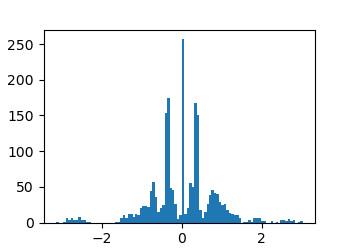}
    \caption{X:56, Y:100}
    \label{fig:init1e}
    \end{subfigure}%
    \begin{subfigure}{0.16\linewidth}
    \includegraphics[scale=0.28]{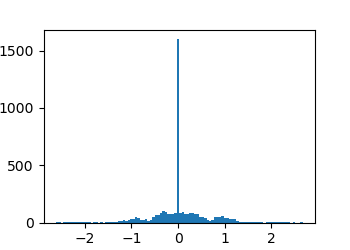}
    \caption{X:110, Y:100}
    \label{fig:init1f}
    \end{subfigure}
    
    \begin{subfigure}{0.16\linewidth}
    \includegraphics[scale=0.28]{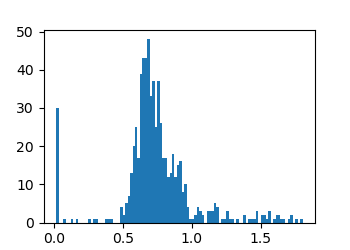}
    {\caption{X:20, Y:10}}
    \label{fig:init2a}
    \end{subfigure}%
    \begin{subfigure}{0.16\linewidth}
    \includegraphics[scale=0.28]{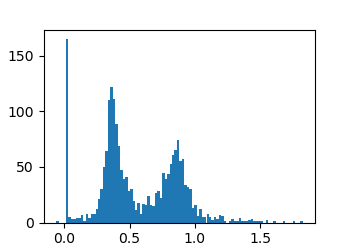}
    \caption{X:56, Y:10}
    \label{fig:init2b}
    \end{subfigure}%
    \begin{subfigure}{0.16\linewidth}
    \includegraphics[scale=0.28]{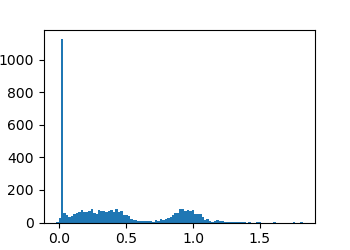}
    \caption{X:110, Y:10}
    \label{fig:init2c}
    \end{subfigure}%
    \begin{subfigure}{0.16\linewidth}
    \includegraphics[scale=0.28]{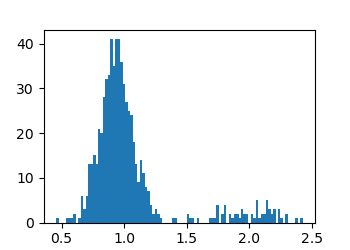}
    \caption{X:20, Y:100}
    \label{fig:init2d}
    \end{subfigure}%
    \begin{subfigure}{0.16\linewidth}
    \includegraphics[scale=0.28]{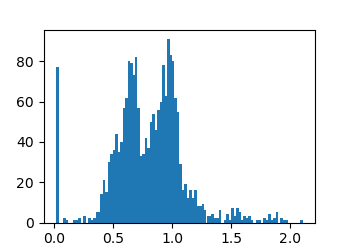}
    \caption{X:56, Y:100}
    \label{fig:init2e}
    \end{subfigure}%
    \begin{subfigure}{0.16\linewidth}
    \includegraphics[scale=0.28]{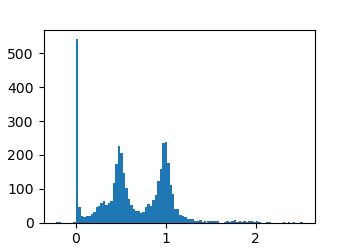}
    \caption{X:110, Y:100}
    \label{fig:init2f}
    \end{subfigure}
    \caption{Change in distribution of slopes for ResNet-X architectures with dataset CIFAR-Y for (i) \textbf{Type-I - top row:} architecture of RReLU is trained from scratch and (ii) \textbf{Type-II - bottom row:} architecture of RReLU whose ReLU slopes are initialized with $1$ of ReLU activation and filter values are initialized with the final filter values obtained from a network with simple ReLU activation. The histplots of the slopes of CIFAR-10, CIFAR-100 and SVHN datasets with WideResNets are given in Fig. \ref{fig:diffinit_extra} in the Appendix.}
    \label{fig:diffinit}
\end{figure}

For some applications, the performance is essential and cannot be sacrificed for the sake of achieving a saving in memory and computation. For these cases where a deeper architecture is used and Type-I initialization faces convergence issues (with CIFAR-10 and CIFAR-100, ResNet-110 archives great saving at the cost of reduced accuracy), we propose to initialize the filters of the RReLU network with the filters of a pre-trained ReLU network and initialize the slopes with simply $+1$. Then the architecture with RReLU is retrained with trainable weights and slopes. This type of initialization has a stable training process and better convergence in the case of deeper architectures at the cost of slightly reduced savings in memory and computation for deeper architectures like ResNet-110. However, with Type-I, more of the slopes go to zero resulting in a less complex network compared to Type-II. The distribution of the slopes $\mathbf{b}_l \forall l\in L$ when different architectures with CIFAR-10 and CIFAR-100 are trained from scratch (Type I - top row) and when they are retrained for $300$ epochs from a pre-trained architecture with ReLU (Type II - bottom row) are shown in Fig. \ref{fig:diffinit}. For the CIFAR-100 dataset, by comparing Fig. \ref{fig:init1e} and \ref{fig:init2e} for ResNet-56 or by comparing Fig. \ref{fig:init1f} and \ref{fig:init2f} for ResNet-110 one can observe that the saving decreases slightly with Type-II initialization but as seen from Table \ref{tab:RReLUperfcaseII}, the accuracy could be improved from $68.96\%$ to $72.83\%$ and $57.27\%$ to $73.67\%$ respectively. For a not so deeper but a wider architecture like WRN-$40$-$4$, the saving in memory and computation is huge with even better accuracy than baseline.

The type-II initialization gives better accuracy for deeper architectures like ResNet-110 with decent savings in memory and computation. For example, with the CIFAR-100 dataset and ResNet-110 architecture, the Type-II initialization and retraining give an accuracy of $73.67\%$ compared to $74.84\%$ of the baseline with a saving of $22.15\%$ saving in memory and $35.20\%$ saving in computation. For other dataset and architectures like WideResNet, this initialization work pretty well. For example, for the same CIFAR-100 dataset, the WRN-$40$-$4$ achieves an accuracy of $79.49\%$ compared to a baseline of $78.82\%$ with a tremendous saving of $60.81\%$ in memory and $74.16\%$ in computation.

\begin{table}[t]
\footnotesize
 \caption{Performance of RReLU with Type-II initialization. To understand the gain in memory and computation, compare the $2^{nd}$ and $3^{rd}$ row of this table with $2^{nd}$ and $3^{rd}$ row of Table \ref{tab:RReLUperf}.}
    \label{tab:RReLUperfcaseII}
    \centering
    \begin{tabular}{p{3.5cm}p{0.8cm}p{0.8cm}p{0.9cm}p{0.9cm}p{0.9cm}p{0.9cm}p{0.9cm}p{0.9cm}p{0.9cm}}
    \toprule
    Dataset  &\multicolumn{4}{c}{CIFAR-10} & \multicolumn{4}{c}{CIFAR-100} & \multicolumn{1}{c}{SVHN} \\
    \midrule
        Architecture & ResNet-20 & ResNet-56 & ResNet-110 & WRN-$40$-$4$ & ResNet-20 & ResNet-56 & ResNet-110 & WRN-$40$-$4$ & WRN-16-4 \\
        \cmidrule(lr){1-1} \cmidrule(lr){2-5} \cmidrule(lr){6-9} \cmidrule(lr){10-10} 
        Filters ignored ($\%$) & $5.36$  & $9.52$ & $30.55$ & $46.75$ & $0.0$  & $4.22$ & $15.30$& $40.26$ &$29.25$\\
        $\#$Params RReLU (Mn) & \textcolor{blue}{$0.24$} & \textcolor{blue}{$0.73$} & \textcolor{blue}{$0.86$} & \textcolor{blue}{$2.89$} & \textcolor{blue}{$0.27$} & \textcolor{blue}{$0.79$} & \textcolor{blue}{$1.34$} & \textcolor{blue}{$3.64$} & \textcolor{blue}{$1.29$}\\
        FLOPs RReLU (Mn) & \textcolor{blue}{$71.31$} & \textcolor{blue}{$188$} & \textcolor{blue}{$244$} & \textcolor{blue}{$666$} & \textcolor{blue}{$81$} & \textcolor{blue}{$229$} & \textcolor{blue}{$327$} &\textcolor{blue}{$717$} & \textcolor{blue}{$360$}\\
        Acc RReLU post-pruning  & $92.62$ & $94.10$ & $92.75$ & $95.72$ & $70.51$ & $72.83$ & $73.67$ & $79.49$ & $96.75$\\
        \bottomrule
    \end{tabular}
   
\end{table}

From Fig. \ref{fig:init1a}, Fig. \ref{fig:init1b} and Fig. \ref{fig:init1c}, it is clear that, with deeper architecture, more number of filters are redundant i.e. with CIFAR-10 dataset, ResNet-20 has $24$, ResNet-56 has $171$ and ResNet-110 has $1438$ redundant filters. One more interesting thing to notice is that if the same architecture is used for two tasks of different levels of complexity, all the filters are exhaustively used in case of more complicated tasks like CIFAR-100 as shown in Fig. \ref{fig:init1d} because there is no RReLU with a slope close to zero; whereas some filters could be pruned for a simpler dataset like CIFAR-10 as shown in Fig. \ref{fig:init1a} without any degradation in performance because there are indeed some RReLUs with slopes close to zero. So based on the requirement of either accuracy or the savings, one can go for either type-I or type-II initialization. RReLU helps to detect them and finally, those filters can be removed to save memory and computation. We have noticed that for type-I initialization with ResNet-110 architecture, many layers did not even have any filters. So the complete information was passing via the skip connection. In the case of type-II, the number of filters dropped per layer is lesser and thus gives a decent saving in complexity with equivalent performance as a baseline.

\subsection{RReLU as coarse feature extractor}
\label{sec:coarse}
The rotation of each RReLU acts like a representation of the coarse features whereas the filters at every layer detect the finer features. To show this we trained only the slopes $\mathbf{b}_l$, keeping weights initialized with Kaiming He initialization\cite{he2015delving}. Whereas the untrained network with simple ReLU does not learn anything, the ResNet-56 with only the learned slopes achieves an accuracy of $53.29\%$ with the CIFAR-10 dataset. The detailed results are in Table \ref{tab:twostep} in Appendix. The slopes are trained for $500$ epochs for ResNet-20 and ResNet-56 and $600$ epochs for ResNet-110 and the histogram of the slopes of ResNet-56 before and after training are given Fig. \ref{fig:histslopes} in Appendix. In this experiment with ResNet-56, one can remove $15.33\%$ of the filters to achieve an accuracy of $52.75\%$ on the CIFAR-10 dataset with complete random initialization of the filters/weights. So it is clear that the proposed activation RReLU learns the coarse features and forces the architecture to use only those filters that are required to represent these features and drops out the rest of the filters that are not useful in each residual block. 

Once the RReLU is used as a coarse feature extractor, one can drop those RReLU which have nearly zero slopes and prune their corresponding filters. We can then train the remaining weights and retrain the slopes. The network after retraining achieves similar accuracy to architecture with ReLU. In this way, saving in complexity is more for ResNet-56 with CIFAR-10 than type-I and type-II as seen in Table \ref{tab:twostep} in the Appendix. However for a complicated task, one should be careful regarding the number of RReLUs dropped after the first step. If more number of RReLUs are dropped in the process, the overall accuracy may not reach the same level as baseline accuracy. The final performance, memory, and computing saving are tabulated in Table \ref{tab:twostep} in Appendix. This way of removing coarse features that are not useful helps to save even more with competitive accuracy compared to Table \ref{tab:RReLUperf}.

\subsection{Features choose to pass via the least number of filters possible}
In \cite{veit2016residual}, the authors have shown residual networks as an ensemble of many independent paths. Very deep residual networks are enabled by only the shorter paths during the training as the longer paths do not contribute to the gradient. With a lesion study they showed that even though one takes deeper architecture for better representation, only the shorter paths carry gradient through them. This implies that the features try to find out shorter paths even though a deeper architecture is used and long paths are present. Our study supports these findings as we observe that features not only try to pass through shorter paths but it tries to pass through a lesser number of filters as well. To study this, we introduce a metric called \textit{filter-path length}. Whereas path length meant the number of layers the input features are passing through, filter-path length means the number of filters the features are passing through. If we look at one residual block, the input can either pass through the skip connection resulting in a filter path length equal to zero, or the weighted path giving rise to a filter-path length of a total number of filters active in the weighted path. For a WRN-$40$-$4$ with ReLU activation, even though the maximum filter path length could be $1424$, because of the structure of ResNets, most of the paths have a filter-path length between $300$ to $1100$ as shown in Fig. \ref{fig:filterpath} in the Appendix. By using RReLU, we observe that the filter-path length reduces i.e. most of the paths have filter-path lengths between a comparatively lower range of $100$ to $500$. Therefore it is clear that the proposed RReLU activation gives the features more freedom to choose to pass through a lesser number of filters and the unnecessary filters can be dropped out resulting in saving memory and computation.

\subsection{RReLU learns to DropOut to achieve better generalization}
\label{sec:dropout}
Dropout \cite{srivastava2014dropout} is a technique that helps the neural networks not to overfit and improves generalization by randomly dropping a set of neurons during training. Instead of dropping out randomly, RReLU learns to drop neurons based on their importance and we term it as \textit{learning to dropout}. The gain in performance with RReLU indicates a better generalization by learning how to drop. For example with the CIFAR-10 dataset, by using proposed RReLU activation, the performance of ResNet-20 improves from $91.25$ to $93.12$ and the performance of ResNet-56 improves from $93.03$ to $93.74$ as shown in Table \ref{tab:RReLUperfdropout} given in the Appendix. This is consistent with WideResNets as well.

\section{Conclusion}
Just by rotating the ReLU, the standard architectures get a new degree of freedom that allows the optimizer to inactivate the filters that are not necessary by tuning the slopes of the RReLU activation. It not only saves the resource but also improves performance in many architectures. For example, with the CIFAR-100 dataset, the WRN-$40$-$4$ has a huge saving of $60.72\%$ in memory requirement and $74.33\%$ in computation without any loss in accuracy. The proposed RReLU can be used as an alternative to ReLU in all those deep architectures where ReLU has been used earlier. It provides a performance that is close to and sometimes better than ReLU while giving a reduction in both computation and memory requirements. We believe that the proposed method can be investigated with recently designed highly efficient EfficientNet family \cite{tan2019efficientnet, foret2020sharpness} or transformer based networks\cite{ridnik2021imagenet} as well and can be interesting future work. The other aspects of neural networks like the adversarial robustness of the proposed RReLU can be tested and can be a potential area to look at.

\bibliographystyle{IEEEtran}
\bibliography{library.bib}

\begin{thebibliography}{10}
\providecommand{\url}[1]{#1}
\csname url@samestyle\endcsname
\providecommand{\newblock}{\relax}
\providecommand{\bibinfo}[2]{#2}
\providecommand{\BIBentrySTDinterwordspacing}{\spaceskip=0pt\relax}
\providecommand{\BIBentryALTinterwordstretchfactor}{4}
\providecommand{\BIBentryALTinterwordspacing}{\spaceskip=\fontdimen2\font plus
\BIBentryALTinterwordstretchfactor\fontdimen3\font minus
  \fontdimen4\font\relax}
\providecommand{\BIBforeignlanguage}[2]{{%
\expandafter\ifx\csname l@#1\endcsname\relax
\typeout{** WARNING: IEEEtran.bst: No hyphenation pattern has been}%
\typeout{** loaded for the language `#1'. Using the pattern for}%
\typeout{** the default language instead.}%
\else
\language=\csname l@#1\endcsname
\fi
#2}}
\providecommand{\BIBdecl}{\relax}
\BIBdecl

\bibitem{krizhevsky2012imagenet}
A.~Krizhevsky, I.~Sutskever, and G.~E. Hinton, ``Imagenet classification with
  deep convolutional neural networks,'' \emph{Advances in neural information
  processing systems}, vol.~25, 2012.

\bibitem{simonyan2014very}
K.~Simonyan and A.~Zisserman, ``Very deep convolutional networks for
  large-scale image recognition,'' \emph{arXiv preprint arXiv:1409.1556}, 2014.

\bibitem{huang2017densely}
G.~Huang, Z.~Liu, L.~Van Der~Maaten, and K.~Q. Weinberger, ``Densely connected
  convolutional networks,'' in \emph{Proceedings of the IEEE conference on
  computer vision and pattern recognition}, 2017, pp. 4700--4708.

\bibitem{szegedy2015going}
C.~Szegedy, W.~Liu, Y.~Jia, P.~Sermanet, S.~Reed, D.~Anguelov, D.~Erhan,
  V.~Vanhoucke, and A.~Rabinovich, ``Going deeper with convolutions,'' in
  \emph{Proceedings of the IEEE conference on computer vision and pattern
  recognition}, 2015, pp. 1--9.

\bibitem{he2016deep}
K.~He, X.~Zhang, S.~Ren, and J.~Sun, ``Deep residual learning for image
  recognition,'' in \emph{Proceedings of the IEEE conference on computer vision
  and pattern recognition}, 2016, pp. 770--778.

\bibitem{zagoruyko2016wide}
S.~Zagoruyko and N.~Komodakis, ``Wide residual networks,'' \emph{arXiv preprint
  arXiv:1605.07146}, 2016.

\bibitem{han2015deep}
S.~Han, H.~Mao, and W.~J. Dally, ``Deep compression: Compressing deep neural
  networks with pruning, trained quantization and huffman coding,'' \emph{arXiv
  preprint arXiv:1510.00149}, 2015.

\bibitem{han2015learning}
S.~Han, J.~Pool, J.~Tran, and W.~Dally, ``Learning both weights and connections
  for efficient neural network,'' \emph{Advances in neural information
  processing systems}, vol.~28, 2015.

\bibitem{wen2016learning}
W.~Wen, C.~Wu, Y.~Wang, Y.~Chen, and H.~Li, ``Learning structured sparsity in
  deep neural networks,'' \emph{Advances in neural information processing
  systems}, vol.~29, 2016.

\bibitem{meier2008group}
L.~Meier, S.~Van De~Geer, and P.~B{\"u}hlmann, ``The group lasso for logistic
  regression,'' \emph{Journal of the Royal Statistical Society: Series B
  (Statistical Methodology)}, vol.~70, no.~1, pp. 53--71, 2008.

\bibitem{wang2017novel}
J.~Wang, C.~Xu, X.~Yang, and J.~M. Zurada, ``A novel pruning algorithm for
  smoothing feedforward neural networks based on group lasso method,''
  \emph{IEEE transactions on neural networks and learning systems}, vol.~29,
  no.~5, pp. 2012--2024, 2017.

\bibitem{nayak2020green}
N.~Nayak, T.~Tholeti, M.~Srinivasan, and S.~Kalyani, ``Green detnet:
  Computation and memory efficient detnet using smart compression and
  training,'' \emph{arXiv preprint arXiv:2003.09446}, 2020.

\bibitem{nair2010rectified}
V.~Nair and G.~E. Hinton, ``Rectified linear units improve restricted boltzmann
  machines,'' in \emph{Icml}, 2010.

\bibitem{maas2013rectifier}
A.~L. Maas, A.~Y. Hannun, A.~Y. Ng \emph{et~al.}, ``Rectifier nonlinearities
  improve neural network acoustic models,'' in \emph{Proc. icml}, vol.~30,
  no.~1.\hskip 1em plus 0.5em minus 0.4em\relax Citeseer, 2013, p.~3.

\bibitem{he2015delving}
K.~He, X.~Zhang, S.~Ren, and J.~Sun, ``Delving deep into rectifiers: Surpassing
  human-level performance on imagenet classification,'' in \emph{Proceedings of
  the IEEE international conference on computer vision}, 2015, pp. 1026--1034.

\bibitem{srivastava2014dropout}
N.~Srivastava, G.~Hinton, A.~Krizhevsky, I.~Sutskever, and R.~Salakhutdinov,
  ``Dropout: a simple way to prevent neural networks from overfitting,''
  \emph{The journal of machine learning research}, vol.~15, no.~1, pp.
  1929--1958, 2014.

\bibitem{guo2020gluoncv}
J.~Guo, H.~He, T.~He, L.~Lausen, M.~Li, H.~Lin, X.~Shi, C.~Wang, J.~Xie, S.~Zha
  \emph{et~al.}, ``Gluoncv and gluonnlp: deep learning in computer vision and
  natural language processing.'' \emph{J. Mach. Learn. Res.}, vol.~21, no.~23,
  pp. 1--7, 2020.

\bibitem{nguyen2021secure}
G.~N. Nguyen, N.~H. Le~Viet, M.~Elhoseny, K.~Shankar, B.~Gupta, and A.~A. Abd
  El-Latif, ``Secure blockchain enabled cyber--physical systems in healthcare
  using deep belief network with resnet model,'' \emph{Journal of Parallel and
  Distributed Computing}, vol. 153, pp. 150--160, 2021.

\bibitem{alrabeiah2020millimeter}
M.~Alrabeiah, A.~Hredzak, and A.~Alkhateeb, ``Millimeter wave base stations
  with cameras: Vision-aided beam and blockage prediction,'' in \emph{2020 IEEE
  91st vehicular technology conference (VTC2020-Spring)}.\hskip 1em plus 0.5em
  minus 0.4em\relax IEEE, 2020, pp. 1--5.

\bibitem{manas2021seasonal}
O.~Ma{\~n}as, A.~Lacoste, X.~Giro-i Nieto, D.~Vazquez, and P.~Rodriguez,
  ``Seasonal contrast: Unsupervised pre-training from uncurated remote sensing
  data,'' in \emph{Proceedings of the IEEE/CVF International Conference on
  Computer Vision}, 2021, pp. 9414--9423.

\bibitem{shankar2021binarized}
N.~P. Shankar, D.~Sadhukhan, N.~Nayak, and S.~Kalyani, ``Binarized resnet:
  Enabling automatic modulation classification at the resource-constrained
  edge,'' \emph{arXiv preprint arXiv:2110.14357}, 2021.

\bibitem{glorot2011deep}
X.~Glorot, A.~Bordes, and Y.~Bengio, ``Deep sparse rectifier neural networks,''
  in \emph{Proceedings of the fourteenth international conference on artificial
  intelligence and statistics}.\hskip 1em plus 0.5em minus 0.4em\relax JMLR
  Workshop and Conference Proceedings, 2011, pp. 315--323.

\bibitem{cybenko1989approximation}
G.~Cybenko, ``Approximation by superpositions of a sigmoidal function,''
  \emph{Mathematics of control, signals and systems}, vol.~2, no.~4, pp.
  303--314, 1989.

\bibitem{barron1993universal}
A.~R. Barron, ``Universal approximation bounds for superpositions of a
  sigmoidal function,'' \emph{IEEE Transactions on Information theory},
  vol.~39, no.~3, pp. 930--945, 1993.

\bibitem{huang2020relu}
C.~Huang, ``Relu networks are universal approximators via piecewise linear or
  constant functions,'' \emph{Neural Computation}, vol.~32, no.~11, pp.
  2249--2278, 2020.

\bibitem{hanin2019universal}
B.~Hanin, ``Universal function approximation by deep neural nets with bounded
  width and relu activations,'' \emph{Mathematics}, vol.~7, no.~10, p. 992,
  2019.

\bibitem{vikas}
D.~Vikas, N.~Nayak, and S.~Kalyani, ``Realizing neural decoder at the edge with
  ensembled bnn,'' \emph{IEEE Communications Letters}, pp. 1--1, 2021.

\bibitem{veit2016residual}
A.~Veit, M.~J. Wilber, and S.~Belongie, ``Residual networks behave like
  ensembles of relatively shallow networks,'' \emph{Advances in neural
  information processing systems}, vol.~29, 2016.

\bibitem{tan2019efficientnet}
M.~Tan and Q.~Le, ``Efficientnet: Rethinking model scaling for convolutional
  neural networks,'' in \emph{International conference on machine
  learning}.\hskip 1em plus 0.5em minus 0.4em\relax PMLR, 2019, pp. 6105--6114.

\bibitem{foret2020sharpness}
P.~Foret, A.~Kleiner, H.~Mobahi, and B.~Neyshabur, ``Sharpness-aware
  minimization for efficiently improving generalization,'' \emph{arXiv preprint
  arXiv:2010.01412}, 2020.

\bibitem{ridnik2021imagenet}
T.~Ridnik, E.~Ben-Baruch, A.~Noy, and L.~Zelnik-Manor, ``Imagenet-21k
  pretraining for the masses,'' \emph{arXiv preprint arXiv:2104.10972}, 2021.

\end{thebibliography}

\appendix

\section{Appendix}
\subsection{Truncated Gaussian Mixture Model for the initialization of the RReLU slopes}
\begin{figure}[h]
    \centering
    \includegraphics[scale=0.6]{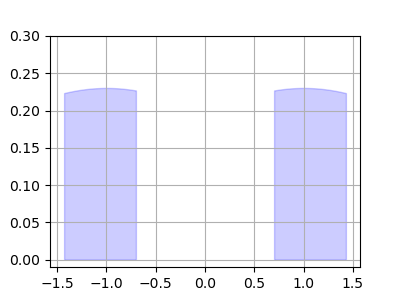}
    \caption{The slope of RReLU are initialized with the values from truncated gaussian mixture model distribution with mean at $\{+1,-1\}$ and variance $3$.}
    \label{fig:gmm}
\end{figure}

The slopes $\mathbf{b}_l$ are neither initialized to close to zero nor too close to infinity so that the learning in stabilized. The range where $\mathbf{b}_l$ is initialized is given in Fig. \ref{fig:gmm}. The Gaussian distributions with mean $+1$ and $-1$ are truncated between $[\tan(35^{\circ}), \tan(55^{\circ})]$ and between for $[\tan(-55^{\circ}), \tan(-35^{\circ})]$ respectively to achieve this. 

\subsection{The sample ResNet and WideResNet architectures with RReLU considered for our experiments}
The architecture of vanilla ResNet-20 is given in Table. \ref{tab:resnet20} where each of the three blocks has $6$ layers. For vanilla ResNet-56 and vanilla ResNet-110, each block has $9$ and $18$ blocks respectively. To get the architecture with RReLU, the ReLU activations at all the layers (except the first convolution layer) are converted to RReLU. We did not convert ReLU to RReLU and kept it as it is at the first convolution layer because we may not want to miss out on any information in the input by deactivating filters at the very first layer. For CIFAR-10 dataset, $K=10$ and for CIFAR-100, $K=100$.  

\begin{table}[h]
\caption{Architecture of vanilla ResNet-20; For the RReLu version, the ReLU activations are replaced by ReLU. Here addition implies the addition of skip connection with the residual path.}
\label{tab:resnet20}
\centering
\footnotesize 
\begin{tabular}{ p{2cm} p{9cm}  p{2.5cm} }
 \toprule
 \textbf{Layer}  & \textbf{ Description} & \textbf{Output Dimension}   \\
 \cmidrule(lr){1-1}\cmidrule(lr){2-2}\cmidrule(lr){3-3}
 Input & - & $3\times32\times32$ \\ 
 Conv & $C_{in}=3, C_{out}=16, $ Kernel: $(3\times3)$, BN, ReLU & $16\times32\times32$ \\
 Block1-layer1 & $C_{in}=16, C_{out}=16$, Kernel: $(3\times3)]$, BN, ReLU & $16\times32\times32$\\
 Block1-layer2 & $C_{in}=16, C_{out}=16$, Kernel: $(3\times3)]$, BN, ReLU, addition & $16\times32\times32$ \\
 Block1-layer3 & $C_{in}=16, C_{out}=16$, Kernel: $(3\times3)]$, BN, ReLU & $16\times32\times32$ \\
 Block1-layer4 & $C_{in}=16, C_{out}=16$, Kernel: $(3\times3)]$, BN, ReLU, addition & $16\times32\times32$ \\
 Block1-layer5 & $C_{in}=16, C_{out}=16$, Kernel: $(3\times3)]$, BN, ReLU & $16\times32\times32$ \\
 Block1-layer6 & $C_{in}=16, C_{out}=16$, Kernel: $(3\times3)]$, BN, ReLU, addition & $16\times32\times32$ \\
 Block2-layer1 & $C_{in}=16, C_{out}=32$, Kernel: $(3\times3)]$, BN, ReLU & $32\times16\times16$\\
 Block2-layer2 & $C_{in}=32, C_{out}=32$, Kernel: $(3\times3)]$, BN, ReLU, addition & $32\times16\times16$ \\
 Block2-layer3 & $C_{in}=32, C_{out}=32$, Kernel: $(3\times3)]$, BN, ReLU & $32\times16\times16$ \\
 Block2-layer4 & $C_{in}=32, C_{out}=32$, Kernel: $(3\times3)]$, BN, ReLU, addition & $32\times16\times16$ \\
 Block2-layer5 & $C_{in}=32, C_{out}=32$, Kernel: $(3\times3)]$, BN, ReLU & $32\times16\times16$ \\
 Block2-layer6 & $C_{in}=32, C_{out}=32$, Kernel: $(3\times3)]$, BN, ReLU, addition & $32\times16\times16$ \\
 Block3-layer1 & $C_{in}=32, C_{out}=64$, Kernel: $(3\times3)]$, BN, ReLU & $64\times8\times8$\\
 Block3-layer2 & $C_{in}=64, C_{out}=64$, Kernel: $(3\times3)]$, BN, ReLU, addition & $64\times8\times8$ \\
 Block3-layer3 & $C_{in}=64, C_{out}=64$, Kernel: $(3\times3)]$, BN, ReLU & $64\times8\times8$ \\
 Block3-layer4 & $C_{in}=64, C_{out}=64$, Kernel: $(3\times3)]$, BN, ReLU, addition & $64\times8\times8$ \\
 Block3-layer5 & $C_{in}=64, C_{out}=64$, Kernel: $(3\times3)]$, BN, ReLU & $64\times8\times8$ \\
 Block3-layer6 & $C_{in}=64, C_{out}=64$, Kernel: $(3\times3)]$, BN, ReLU, addition & $64\times8\times8$ \\
 Linear & classification into K classes with softmax & $K\times1$\\
 \bottomrule
\end{tabular}
\end{table}

The WRN-$16$-$4$ architecture is given in Table. \ref{tab:WRN} that is used with SVHN dataset. For SVHN $K=10$. The other WideResNet architecture is WRN-$40$-$4$ that has $40$ layers and widening factor $4$. It has same three blocks but each with $12$ layers.
\begin{table}[h]
\caption{Architecture of WRN-$16$-$4$}
\label{tab:WRN}
\centering
\footnotesize 
\begin{tabular}{ p{2cm} p{9cm} p{2.5cm} }
 \toprule
 \textbf{Layer}  & \textbf{ Description} & \textbf{Output Dimension}   \\
\cmidrule(lr){1-1}\cmidrule(lr){2-2}\cmidrule(lr){3-3}
 Input & - & $3\times32\times32$ \\ 
 Conv & $C_{in}=3, C_{out}=16, $ Kernel: $(3\times3)$, BN, ReLU & $16\times32\times32$ \\
 Block1-layer1 & $C_{in}=16, C_{out}=64$, Kernel: $(3\times3)]$, BN, ReLU & $64\times32\times32$\\
 Block1-layer2 & $C_{in}=64, C_{out}=64$, Kernel: $(3\times3)]$, BN, ReLU, addition & $64\times32\times32$ \\
 Block1-layer3 & $C_{in}=64, C_{out}=64$, Kernel: $(3\times3)]$, BN, ReLU & $64\times32\times32$ \\
 Block1-layer4 & $C_{in}=64, C_{out}=64$, Kernel: $(3\times3)]$, BN, ReLU, addition & $64\times32\times32$ \\
 Block2-layer1 & $C_{in}=64, C_{out}=128$, Kernel: $(3\times3)]$, BN, ReLU & $128\times16\times16$\\
 Block2-layer2 & $C_{in}=128, C_{out}=128$, Kernel: $(3\times3)]$, BN, ReLU, addition & $128\times16\times16$ \\
 Block2-layer3 & $C_{in}=128, C_{out}=128$, Kernel: $(3\times3)]$, BN, ReLU & $128\times16\times16$ \\
 Block2-layer4 & $C_{in}=128, C_{out}=128$, Kernel: $(3\times3)]$, BN, ReLU, addition & $128\times16\times16$ \\
 Block3-layer1 & $C_{in}=128, C_{out}=256$, Kernel: $(3\times3)]$, BN, ReLU & $256\times8\times8$\\
 Block3-layer2 & $C_{in}=256, C_{out}=256$, Kernel: $(3\times3)]$, BN, ReLU, addition & $256\times8\times8$ \\
 Block3-layer3 & $C_{in}=256, C_{out}=256$, Kernel: $(3\times3)]$, BN, ReLU & $256\times8\times8$ \\
 Block3-layer4 & $C_{in}=256, C_{out}=256$, Kernel: $(3\times3)]$, BN, ReLU, addition & $256\times8\times8$ \\
 Linear & classification into K classes with softmax & $K\times1$\\
 \bottomrule
\end{tabular}
\end{table}

\subsection{How to set $\gamma$ during deployment of the models to discard the slopes that are below $\gamma$ and still not degrade the performance?}
The value of $\gamma$ is set by cross-validation. The network is trained with the train set and the test set is divided into two parts. The first part of the test set is used to find a $\gamma$ such that if the slopes that are below that $\gamma$ are made equal to zero, the performance does not degrade. The number of RReLU slopes that can be ignored for different experiments and the corresponding $\gamma$s is given in Table \ref{tab:RReLUperfgamma}. With the latter part of the test set, the pruned network is tested and accuracy is reported in Table \ref{tab:RReLUperf}. 
\begin{table}[h]
    \caption{The value of $\gamma$ for different experiments and the corresponding number of filter ignored in `x/y' format. Here `x' filters out of `y' filters could be dropped because of RReLU. $1^{st}$ two rows and $2^{nd}$ two rows are for experimnts with Type-I (Sec. \ref{sec:type-I}) and Type-II (Sec. \ref{sec:smarterinit}) initialization respectively and $3^{rd}$ two rows are for coarse feature extraction and retraining (Sec. \ref{sec:coarse})}
    \label{tab:RReLUperfgamma}
    \centering
    \footnotesize 	
    \begin{tabular}{p{1.2cm}p{0.5cm}p{0.7cm}p{0.7cm}p{0.7cm}p{0.7cm}p{0.7cm}p{0.7cm}p{0.7cm}p{0.7cm}p{0.7cm}}
    \toprule
    Dataset &\multicolumn{1}{c}{MNIST} &\multicolumn{4}{c}{CIFAR-10} & \multicolumn{4}{c}{CIFAR-100} &\multicolumn{1}{c}{SVHN}\\
    \cmidrule(lr){1-1} \cmidrule(lr){2-2} \cmidrule(lr){3-6} \cmidrule(lr){7-10} \cmidrule(lr){11-11} 
        Architecture & FCNN & ResNet-20 & ResNet-56 & ResNet-110 & WRN-40-4 & ResNet-20 & ResNet-56 & ResNet-110 & WRN-40-4 & WRN-16-4\\
        \midrule
        $\gamma$ & $1.0$ &  $0.2$ & $0.06$ & $0.06$ & $0.045$ & $0$ & $0.16$ & $0.02$ & $0.03$ & $0.04$ \\
        Filters ignored & $\frac{24}{500}$ & $\frac{15}{672}$ & $\frac{171}{2016}$ & $\frac{1438}{4032}$ & $\frac{2262}{5136}$   & $\frac{0}{672}$ & $\frac{309}{2016}$ & $\frac{1587}{4032}$ &  $\frac{940}{5136}$ & $\frac{324}{1552}$\\
        \midrule
        $\gamma$ & - & $0.38$ & $0.15$ & $0.06$ & $0.05$ &  $0$ &  $0.25$ & $0.1$ & $0.04$ & $0.06$\\
        Filters ignored & - & $\frac{36}{672}$  & $\frac{192}{2016}$ & $\frac{1232}{4032}$ & $\frac{2401}{5136}$  & $\frac{0}{672}$ & $\frac{85}{2016}$ & $\frac{617}{4032}$  & $\frac{2068}{5136}$  & $\frac{454}{1552}$\\
        \midrule
        $\gamma$ & - & $0.04$ & $0.1$ & - & - & $0.03$ & $0.11$ & - & - & - \\
        Filters ignored & - & $\frac{37}{672}$ & $\frac{309}{2016}$ & - & - & $\frac{39}{672}$ & $\frac{373}{2016}$ & - & - & -  \\
        \bottomrule
    \end{tabular}
\end{table}

\subsection{Comparison of the distribution of the slopes of RReLU between type-I and type-II initialization for WideResNets}
\begin{figure}[h]
    \centering
    \begin{subfigure}{0.33\linewidth}
    \includegraphics[scale=0.55]{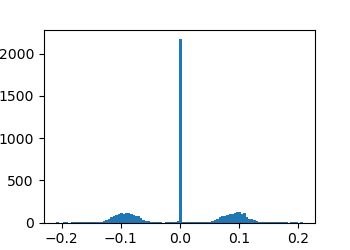}
    \caption{WRN-40-4, CIFAR-10}
    \label{fig:init1a_extra}
    \end{subfigure}%
    \begin{subfigure}{0.33\linewidth}
    \includegraphics[scale=0.55]{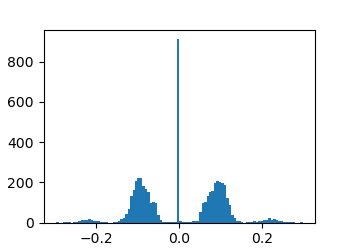}
    \caption{WRN-40-4, CIFAR-100}
    \label{fig:init1b_extra}
    \end{subfigure}%
    \begin{subfigure}{0.33\linewidth}
    \includegraphics[scale=0.55]{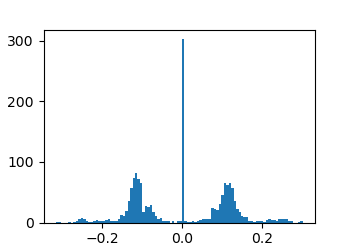}
    \caption{WRN-16-4, SVHN}
    \label{fig:init1c_extra}
    \end{subfigure}
    \begin{subfigure}{0.33\linewidth}
    \includegraphics[scale=0.55]{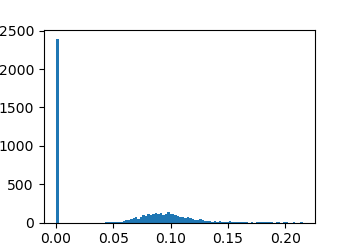}
    \caption{WRN-40-4, CIFAR-10}
    \label{fig:init2a_extra}
    \end{subfigure}%
    \begin{subfigure}{0.33\linewidth}
    \includegraphics[scale=0.55]{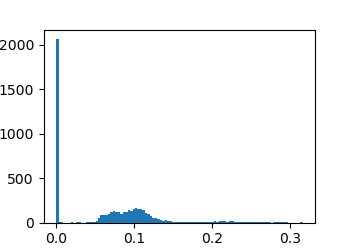}
    \caption{WRN-40-4, CIFAR-100}
    \label{fig:init2b_extra}
    \end{subfigure}%
    \begin{subfigure}{0.33\linewidth}
    \includegraphics[scale=0.55]{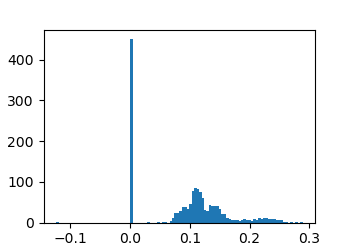}
    \caption{WRN-16-4, SVHN}
    \label{fig:init2c_extra}
    \end{subfigure}%
    \caption{Change in distribution of slopes for \textbf{(a, d)} architecture WRN-$40$-$4$ with dataset CIFAR-10, \textbf{(b, e)} architecture WRN-$40$-$4$ with dataset CIFAR-100 and \textbf{(c, f)} architecture WRN-$16$-$4$ with dataset SVHN for (i) \textbf{Type-I - top row:} architecture of RReLU is trained from scratch and (ii) \textbf{Type-II - bottom row:} architecture of RReLU whose slopes are initialized with $1$ of ReLU activation and filter values are initialized with the final filter values obtained from a network with simple ReLU activation}
    \label{fig:diffinit_extra}
\end{figure}

In the main manuscript, the distributions of the slope of RReLU are shown for ResNets for CIFAR-10 and CIFAR-100 datasets. The same with WRN-16-4 and WRN-40-4 architectures are shown in Fig. \ref{fig:diffinit_extra}. It is clear in the figure that many slopes take values close to zero indicating that the corresponding filters are not useful. For WideResNets, a great number of filters could be ignored without any degradation in performance (For accuracy and saving in memory and computation, see Table \ref{tab:RReLUperf}). In the case of WideResNets, more filters could be ignored if type-II initialization is used i.e. the trainable RReLU slopes are initialized with $1$ and the filters are initialized with the trained filters of the architecture with ReLU.

\begin{table}[h]
    \centering
    \caption{Performance of RReLU with two step training where the first step is to train only the slopes keeping weights (initialized by Kaiming He initialization) non-trainable to extract the coarse features and second step is to retrain the weights and the slopes together.}
    \label{tab:twostep}
    \footnotesize 
    \begin{tabular}{p{4cm}p{1.5cm}p{1.5cm}p{1.5cm}p{1.5cm}}
    \toprule
    Dataset &\multicolumn{2}{c}{CIFAR-10} & \multicolumn{2}{c}{CIFAR-100} \\
    \cmidrule(lr){1-1} \cmidrule(lr){2-3} \cmidrule(lr){4-5} 
        Architecture & ResNet-20 & ResNet-56 & ResNet-20 & ResNet-56  \\
        \midrule
        Acc ReLU & $91.25$ & $93.03$ & $68.20$ & $69.99$ \\
        $\#$Params ReLU &  \textcolor{red}{$0.27$} & \textcolor{red}{$0.85$} &  \textcolor{red}{$0.27$} & \textcolor{red}{$0.85$} \\
        FLOPs ReLU &  \textcolor{red}{$81.19$} & \textcolor{red}{$231.47$} &  \textcolor{red}{$81.20$} & \textcolor{red}{$231.48$} \\
        \midrule
        Acc RReLU (trainable $\mathbf{b}_l$) & $44.69$ & $53.29$ &  $7.96$ & $9.03$  \\
        Filters ignored ($\%$) & $5.5$ & $15.33$ &  $5.8$ & $18.5$  \\
        $\#$Params RReLU & \textcolor{blue}{$0.24$} & \textcolor{blue}{$0.59$} & \textcolor{blue}{$0.25$} & \textcolor{blue}{$0.55$}  \\
        FLOPs RReLU & \textcolor{blue}{$74.57$} & \textcolor{blue}{$185.72$} & \textcolor{blue}{$72.58$} & \textcolor{blue}{$172.14$}  \\
        Acc RReLU & $92.20$ & $93.34$ & $66.61$ & $67.81$  \\
        \bottomrule
    \end{tabular}
\end{table}

\begin{figure}[h]
    \centering
    \begin{subfigure}{0.33\linewidth}
    \includegraphics[scale=0.5]{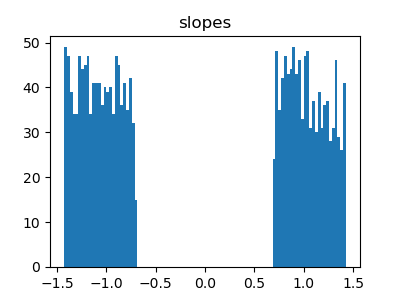}
    \caption{Before training}
    \end{subfigure}%
    \begin{subfigure}{0.33\linewidth}
    \includegraphics[scale=0.5]{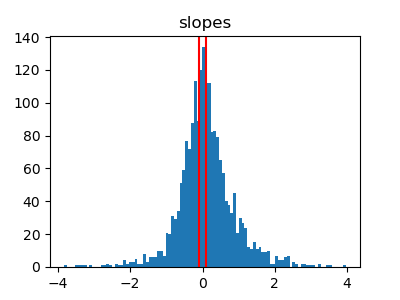}
    \caption{After training for CIFAR-10}
    \end{subfigure}%
    \begin{subfigure}{0.33\linewidth}
    \includegraphics[scale=0.5]{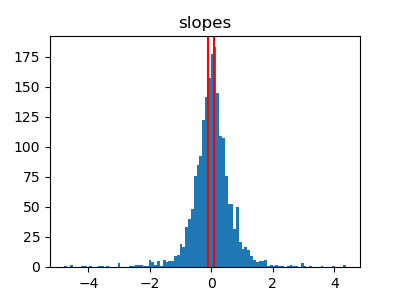}
    \caption{After training for CIFAR-100}
    \end{subfigure}
    \caption{Distribution of a total of $2016$ slopes of RReLU for ResNet-56. (a) Initial distribution of slopes with truncated GMM. For CIFAR-10, (b) $309$ out of $2016$ and for CIFAR-100, (c) $373$ out of $2016$ slopes are less than $|0.1|$ and $|0.11|$ respectively.}
    \label{fig:histslopes}
\end{figure}
\subsection{Results on two-step training: discarding unnecessary coarse features and retraining}
The results in Table \ref{tab:twostep} and Fig. \ref{fig:histslopes} correspond to Subsec. \ref{sec:coarse} where the architecture with RReLU is trained in two steps. At first, the model is trained by keeping only the slopes trainable but the filters non-trainable. Once the coarse features are extracted by the RReLU slopes, the unnecessary coarse features are discarded. At the next step, both slopes and the filters are trained to achieve an accuracy as good as the vanilla ResNets. The fourth row of Table \ref{tab:twostep} gives the accuracy when the architecture with RReLU is trained with only trainable slopes of ReLU but the weights are nontrainable and initialized with uniform random Kaiming He initialization. With all random values in the filters and just the ReLU slopes trained, the models achieve some accuracy which shows that RReLU extracts the coarse features which are later fine-tuned by training the filters along with the slopes. In Fig. \ref{fig:histslopes}, the distribution of slopes is shown when the weights/filters are not trainable. It is clear that for both CIFAR-10 and CIFAR-100 datasets, some RReLU activations have slopes of zero. Once the unnecessary filter are discarded and the rest of the network is retrained, the model achieves accuracy close to the vanilla ResNets with a great saving in memory and computation as shown in Table \ref{tab:twostep}.

\subsection{Results on reduced filter-path length using RReLU}
\begin{figure}[h]
    \centering
    \includegraphics[scale=0.5]{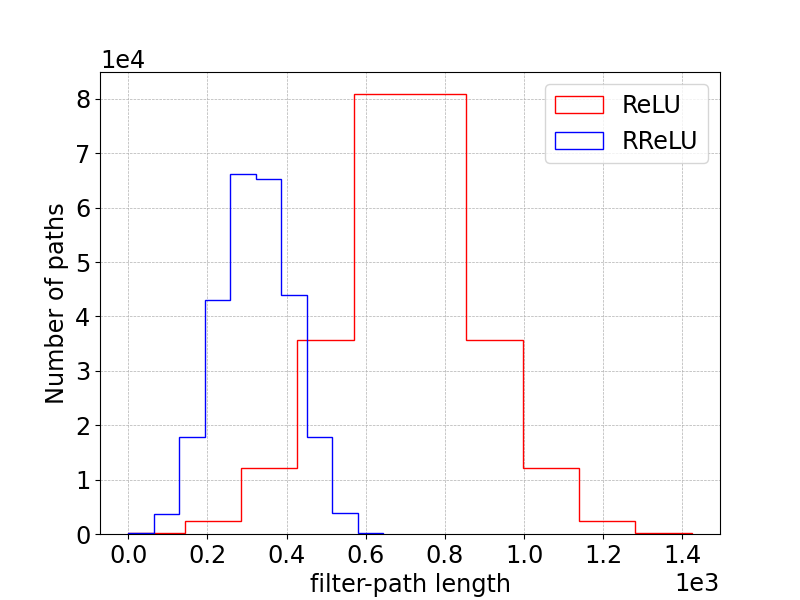}
    \caption{Distribution of filter-path length for WRN-$40$-$4$ when trained with CIFAR-100. The RReLU activation allows the features to pass through less number of filters and therefore with RReLU, the filter-path length decreases.}
    \label{fig:filterpath}
\end{figure}

In Fig. \ref{fig:filterpath}, the distribution of filter-path lengths of architectures with ReLU and RReLU are plotted and it is clear that the filter-path length of architectures with RReLU are smaller than architectures with ReLU. It shows that when the architectures have a choice to discard some filters (like in RReLU), the features choose to pass via a lesser number of filters.  

\subsection{Results on RReLU learns to dropout to achieve better generalization}
As discussed in Subsec. \ref{sec:dropout}, RReLU learns to dropout where the architecture has better representation power or more complexity than the task needs. Therefore RReLU presents a better generalization for those architectures that otherwise overfit. For deeper architectures like ResNet-110, both ReLU and RReLU encounter convergence issues if highly tuned hyperparameters are not chosen and therefore kept aside for this study. Other than ResNet-110, all other architectures that have higher complexity (more parameters than needed) show better generalization with RReLU consistently. Therefore with not-so-deep architectures, the performance improves consistently as highlighted in Table \ref{tab:RReLUperfdropout}. 
\begin{table}[h]
    \caption{Dropout improves the generalization capability of a NN and RReLU learns to dropout some parameters in a structured way. Except for very deep architectures like ResNet-110, the performance improves indicating a better generalization using RReLU. The accuracies reported for architectures with RReLU are with type-I initialization unless stated differently inside the bracket.}
    \label{tab:RReLUperfdropout}
    \centering
    \footnotesize
    \begin{tabular}{p{2.2cm}p{2.0cm}p{2.0cm}p{2.0cm}p{2.0cm}p{1.7cm}}
    \toprule
    Dataset &\multicolumn{1}{c}{MNIST} &\multicolumn{4}{c}{CIFAR-10} \\
    \cmidrule(lr){1-1} \cmidrule(lr){2-2} \cmidrule(lr){3-6} 
        Architecture & FCNN & ResNet-20 & ResNet-56 & ResNet-110 & WRN-40-4 \\
        \midrule
        Acc ReLU & $98.33$ & $91.25$& $93.03$ & $93.57$ & $95.47$ \\
        Acc RReLU & $98.25$ &$\mathbf{93.15}$ & $\mathbf{94.10}$ & $92.75$ (type-II)  & $\mathbf{95.71}$ \\
        \bottomrule
        \vspace{1mm}
    \end{tabular}
    \begin{tabular}{p{2.2cm}p{2.0cm}p{2.0cm}p{2.0cm}p{2.0cm}p{1.7cm}}
    \toprule
    Dataset & \multicolumn{4}{c}{CIFAR-100} &\multicolumn{1}{c}{SVHN}\\
    \cmidrule(lr){1-1} \cmidrule(lr){2-5} \cmidrule(lr){6-6} 
        Architecture & ResNet-20 & ResNet-56 & ResNet-110 & WRN-40-4 & WRN-16-4\\
        \midrule
        Acc ReLU  & $68.20$ & $69.99$& $74.84$ & $78.82$ & $97.01$\\
        Acc RReLU & $\mathbf{70.51}$(type-II) & $\mathbf{72.83}$ (type-II)  & $73.67$ (type-II) & $\mathbf{80.72}$  & $\mathbf{97.053}$\\
        \bottomrule
    \end{tabular}
    
\end{table}

\end{document}